\documentclass[conference]{IEEEtran}
\IEEEoverridecommandlockouts
\usepackage{cite}
\usepackage{amsmath,amssymb,amsfonts}
\usepackage{algorithmic}
\usepackage{graphicx}
\ifCLASSOPTIONcompsoc
\usepackage[caption=false,font=normalsize,labelfont=sf,textfont=sf]{subfig}
\else
\usepackage[caption=false,font=footnotesize]{subfig}
\fi
\usepackage{textcomp}
\usepackage{xcolor}
\usepackage{algorithm}
\usepackage{comment}
\def\BibTeX{{\rm B\kern-.05em{\sc i\kern-.025em b}\kern-.08em
		T\kern-.1667em\lower.7ex\hbox{E}\kern-.125emX}}
\begin{document}
	
	\title{CANS: Communication Limited Camera Network Self-Configuration for Intelligent Industrial Surveillance \\
	}
	
	\author{
		Jingzheng~Tu, Qimin~Xu, Cailian~Chen,~\IEEEmembership{Member,~IEEE}  \\
		\textit{Department of Automation, Shanghai Jiao Tong University}, Shanghai 200240, China \\
		Key Laboratory of System Control and Information Processing, Ministry of Education of China \\ 
		E-mail: \{tujingzheng, qiminxu, cailianchen\}@sjtu.edu.cn
	}
	
	\maketitle
	
	\begin{abstract}
		
		Realtime and intelligent video surveillance via camera networks involve computation-intensive vision detection tasks with massive video data, which is crucial for safety in the edge-enabled industrial Internet of Things (IIoT). 
		Multiple video streams compete for limited communication resources on the link between edge devices and camera networks, resulting in considerable communication congestion. It postpones the completion time and degrades the accuracy of vision detection tasks. Thus, achieving high accuracy of vision detection tasks under the communication constraints and vision task deadline constraints is challenging. 
		Previous works focus on single camera configuration to balance the tradeoff between accuracy and processing time of detection tasks by setting video quality parameters. 	In this paper, an adaptive camera network self-configuration method (CANS) of video surveillance is proposed to cope with multiple video streams of heterogeneous quality of service (QoS) demands for edge-enabled IIoT. Moreover, it adapts to video content and network dynamics. 
		Specifically, the tradeoff between two key performance metrics, \emph{i.e.,} accuracy and latency, is formulated as an NP-hard optimization problem with latency constraints. Simulation on real-world surveillance datasets demonstrates that the proposed CANS method achieves low end-to-end latency (13 ms on average) with high accuracy (92\% on average) with network dynamics. The results validate the effectiveness of the CANS.
	\end{abstract}
	

	\section{Introduction}
	With the development of industrial intelligence, edge-enabled industrial Internet of Things (IIoT) becomes increasingly attractive due to lower response latency and higher resource utilization. 
	Among lots of perception terminals (\emph{e.g.,} thermocouples, thermal imagers, and industrial cameras) deployed in IIoT, industrial camera networks provide copious irreplaceable content information for surveillance safety. 
	Realtime and intelligent video surveillance via industrial camera networks is significant for safety in industrial factories. It requires low end-to-end latency and high accuracy of real-time vision detection tasks. 
	However, video surveillance demands intensive computing resources due to massive high-dimensional video data. 
	Besides, constrained communication resources of camera network surveillance result in high communication latency and high computing costs. 
	Therefore, designing an effective camera network self-configuration method that performs high accuracy of vision detection tasks under communication limitations and latency constraints is vital for intelligent surveillance in edge-enabled IIoT.
	
	
	\begin{figure}[!t]
		\centerline{\includegraphics[width=9cm]{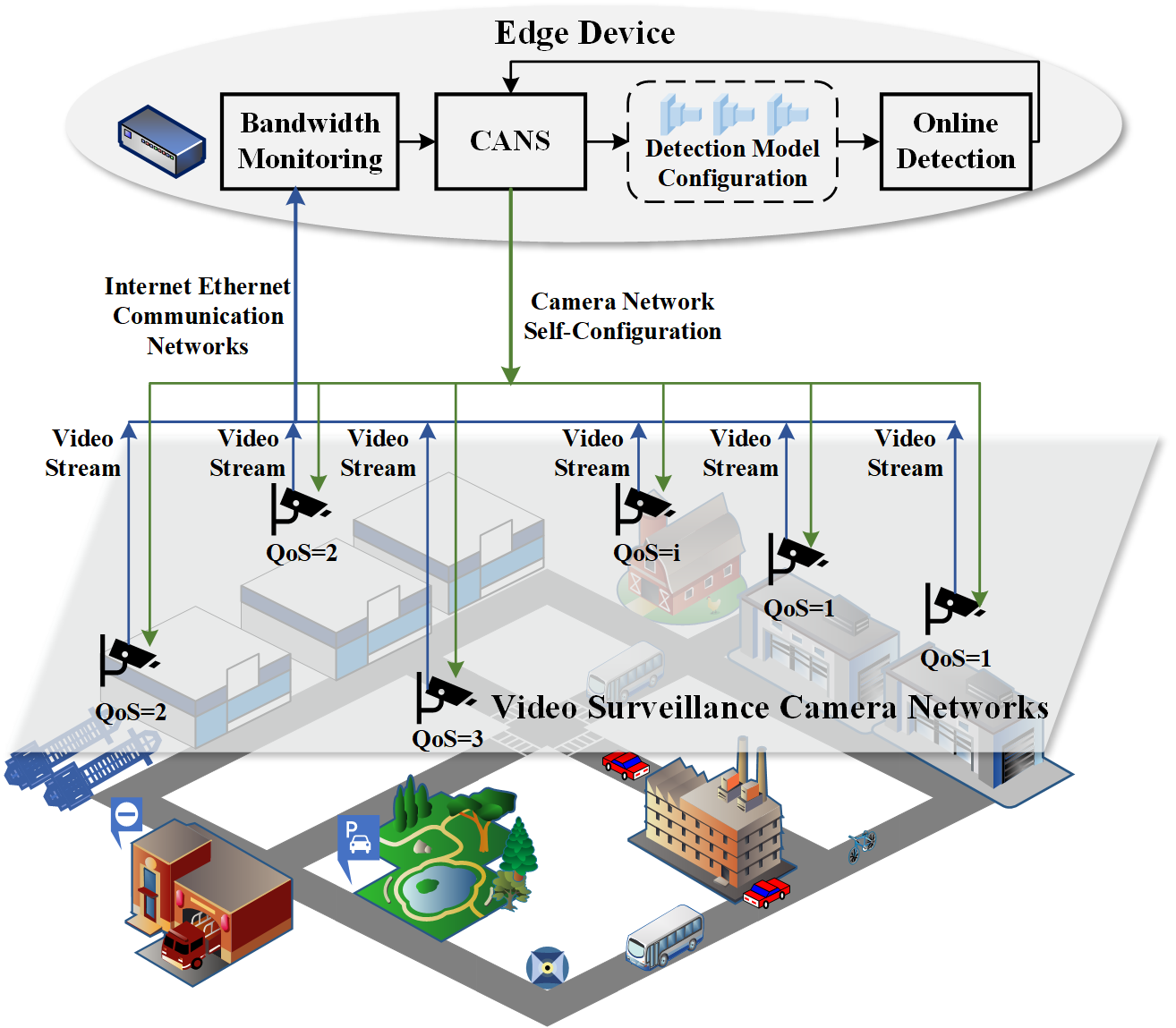}}
		\caption{The framework of the proposed edge-enabled adaptive camera network self-configuration method, CANS. Bandwidth resources are pre-allocated. 		}
		\label{fig:framwork}
		\vspace{-15pt}
	\end{figure}

	A group of works \cite{jiang2018chameleon, wang2020surveiledge} focuses on reducing the demand for computing resources in video surveillance. Concretely, Chameleon \cite{jiang2018chameleon} uses spatial and temporal correlations for periodic video configuration and verifies the dependence of configuration knobs on the accuracy metric. Besides, SurveilEdge \cite{wang2020surveiledge} presents a framework of offline clustering and online finetuning on a cloud center. Then, SurveilEdge deploys the finetuned deep learning networks on each edge node. However, they do not discuss the impact of network resources on key performance metrics of video configuration. 
	Another group of works \cite{fang2018nestdnn, xu2018deepcache, huynh2017deepmon} contributes to the continuous mobile vision. These works propose compressing deep learning models and scheduling computing resources with less accuracy loss on mobile nodes.  
	However, they study computing resource optimization with a single camera instead of a camera network. 
	Meanwhile, DeepDecision \cite{ran2018deepdecision}, JCAB \cite{wang2020joint} and FastVA \cite{tan2020fastva} study the tradeoffs between accuracy, latency and energy costs of video surveillance with multiple video streams. 
	However, they assume the priorities of all input video streams are the same. In practice, different fields of view of cameras in open industrial factories lead to various QoS demands of video streams.

	Therefore, this paper proposes an adaptive camera network self-configuration method (CANS) of video surveillance to deal with multiple video streams of heterogeneous QoS under network dynamics in edge-enabled IIoT. Concretely, we formulate an optimization problem and impose latency constraints and network resource limitations. The objective is to minimize the total end-to-end latency of multiple video streams with different QoS demands while maintaining the high accuracy of vision detection tasks. Besides, an efficient algorithm is proposed to solve the problem. Simulations demonstrate the effectiveness of the proposed CANS. In sum, the main contributions of this paper are as follows:
	\begin{itemize}
		\item An adaptive camera network self-configuration method is proposed to consider video streams with QoS heterogeneity for video surveillance in edge-enabled IIoT, which adapts to network dynamics and video contents. 
		\item An optimization problem is established with latency and network resource constraints. It aims to minimize the end-to-end latency of multiple video streams with heterogeneous QoS while achieving higher accuracy performance.  
		\item An efficient algorithm is presented to solve the above problem. Simulations on real-world surveillance datasets validate the effectiveness of the CANS.
	\end{itemize}

	\section{System Model and Problem Formulation}
	In this section, the overall system model of the proposed CANS method is described in detail, including a video surveillance network model, a latency model, and an accuracy model. Then, the problem formulation is given. 
	
	\subsection{Video Surveillance Network Model}
\begin{figure}[!t]
		\centerline{\includegraphics[width=9cm]{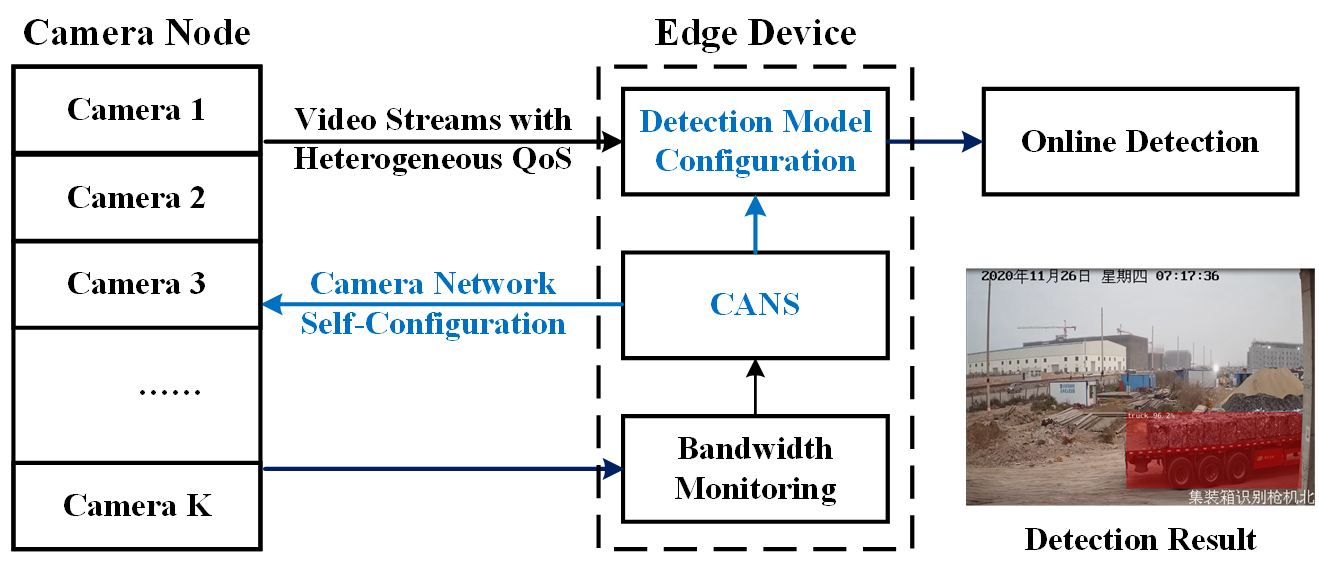}}
		\caption{The architecture of the proposed CANS method.}
		\label{fig:architure}
		\vspace{-15pt}
	\end{figure}

	We mainly focus on vision detection tasks in which targets are operators, mobile trunks, and devices. Once detecting an operator approaching some dangerous areas, a conspicuous alarm is reported instantaneously. Fig. \ref{fig:framwork} shows the framework of the proposed CANS method. The system architecture is shown in Fig. \ref{fig:architure}. Multiple cameras continuously monitor real-time operations in open industrial factories and deliver the captured videos to a nearby edge device. On the edge device, the CANS considers network resource limitation, accuracy, and latency goals to adjust camera network configuration adaptively. Finally, the configuration profile is sent back to the cameras.

	Suppose that a set of $K$ cameras or video streams, denoted by $\mathcal{V}=\{v_1, v_2, \cdots, v_K\}$, connects to one nearby edge device deployed in intelligent factories. These cameras continuously transmit real-time video streams to the edge device with a shared narrow uplink channel. Since these cameras' fields of view are sub-regions with different degrees of importance and danger, the QoS demands of different cameras' video streams possess heterogeneity. The QoS requirements for these $K$ cameras are represented by $\mathcal{Q}=\{q_1, q_2, \cdots, q_K\}$. 
	
	Besides, $N$ parallel convolutional neural network (CNN) models $\mathcal{M}=\{m_1, m_2, \cdots, m_N\}$ are deployed on the edge device with different input sizes of images. We use $r_i$ to denote the input shape or resolution for the $i$-th CNN model. 
	Ref. \cite{simonyan2015very} reports that model compression techniques can diminish the size of a CNN model (\emph{e.g.} removing computing-intensive layers and reducing input shapes) at the expense of accuracy. Thus, we consider that a CNN model with a smaller input resolution has fewer resource requirements, faster processing speed, and lower processing latency.

	We consider the communication network as a cellular network in the proposed CANS method, supporting the D2D communication mode. Cameras in the industrial factory deliver video streams to the edge device via D2D links. The assumption is that all the D2D links share the same channel. Moreover, the transmission rate from the $i$-th camera to the edge device is denoted by $p_i$, defined as the video data transmitted per second, \emph{e.g.}, 1 Mbps. 
	Moreover, the output data of vision detection tasks on the edge device has a much smaller size than the input subsequent video data \cite{long2019edge}. As a result, the transmission time of the edge device on collecting vision detection results is ignored. 
	
	\subsubsection{Latency Model}
	The end-to-end latency $l_{ij}$ per frame of the video $v_i$ and the model $m_j$ is defined as the total latency of one complete image delivered to the edge device and processed by the model $m_j$, which contains the transmission latency and the processing latency. 
	The transmission latency is determined by the frame size and the transmission bandwidth. Suppose the bandwidth is $b$. Thus, the latency expetation per frame of $v_i$ and $m_j$ is written as:
	\begin{equation}
	l_{ij} = \mathbb{E}\{l_{ij}^{\text{tran}} (r_i) \} +  \mathbb{E}\{l_{ij}^{\text{proc}} (r_i, x_{ij})\} = \frac{p_i}{f_i b} + l_{ij}^{\text{CNN}} (r_i, x_{ij}),
	\end{equation}
	where $f_i$ is the framerate of the video $v_i$ and $l_{ij}^{\text{CNN}}$ denotes the processing latency by the CNN model $m_j$. The video bitrate of the video $v_i$ is denoted by $p_i$. The relationship between video bitrate and video resolution is $p_i = \alpha r_i^2$, where $\alpha$ is the bit size per pixel in one video frame. 
	As a result, the latency of video stream $v_i$ per frame is $\sum_{i=1}^{M} l_{ij} x_{ij}$.

	\subsubsection{Accuracy Model of Vision Detection Tasks}
	The analytics accuracy is affected by multiple variables in different ways. These relationships or tradeoffs do not have explicit analytic expressions, leading to complicated solving procedures. For instance, the accuracy depends on the video content, the "black box" of deep learning models, and the video resolution and compression. 
	Thus, establishing a precise accuracy model under a particular configuration is of great challenge. 
	
	The main variables that influence the analytics accuracy (also called configuration knobs) mainly include the input image resolution, the CNN model, and the video stream's bitrate. We consider the impact of the image resolution and the CNN model on analytics accuracy. 
	The relationship between each configuration knob and the accuracy is obtained by real experiments. The accuracy function $a_{ij}(r_i, x_{ij})$ denotes the vision detection accuracy of video $v_i$ with a selected resolution $r_i$ and model $m_j$. A binary variable $x_{ij}$ indicates whether the model $m_j$ is selected for the video $v_i$. Consequently, $\sum_{i=1}^{N} x_{ij}r_i$ is the input frame resolution for video $v_i$.

	\subsection{Problem Formulation}
	The previous section indicates that the relationships between the decision variables and the key performance metrics are complicated. Some variables affect multiple metrics in opposed ways. For instance, higher image resolution results in better accuracy but brings more significant latency. Meanwhile, some variables influence the same metric in different ways. For example, delivering video streams with a higher bitrate increases the latency and decreases the accuracy simultaneously. Thus, selecting a proper combination of decision variables to optimize multiple key metrics is challenging.
	
	Therefore, our objective is to establish an adaptive optimization problem to maximize the vision detection accuracy while minimizing the latency of multiple video streams with heterogeneous QoS requirements with bandwidth limitation and latency constraints. 
	To solve this problem, we consider two key issues: the former is constructing the latency constraints with accuracy guarantees of vision detection tasks; the latter is solving the optimization problem efficiently concerning network dynamics. Hence, these two issues are introduced in the remainder of this work.

	\section{The CANS Algorithm}
	In this section, an optimization problem of the proposed CANS is formulated. Moreover, an efficient algorithm is presented to solve the optimization problem.

	\subsection{Construction of Constraint Set}
	The edge device monitors network resources incessantly. Once the network resources have a significant fluctuation over the predetermined threshold, the CANS determines the camera network configuration for each video stream $v_i$: the video resolution $r_i$ and which model $x_{ij}$ to use.
	
	\subsubsection{Latency Constraints}
	The latency of each video stream must satisfy the minimum tolerable response time:
	\begin{align}
	\label{eq:1}
	\sum\limits_{j=1}^{N} l_{ij} x_{ij} \leq \min\limits_{i} L_i ,   i=1,\ldots,K ,
	\end{align}
	where $L_i$ represents the tolerable latency deadline of video stream $v_i$ when processing vision detection tasks. 
	
	Besides, the CNN processing latency of video stream $v_i$ per frame cannot exceed the reciprocal of video framerate due to the realtime requirement of vision detection. Thus, 
	\begin{align}
	\label{eq:2}
	\sum\limits_{j=1}^N x_{ij} l_{ij}^{\text{CNN}}\leq \frac{1}{f_i},   i=1,\ldots,K.
	\end{align}

	\subsubsection{Bandwidth Constraint}
	The video bitrate summation of all $K$ video streams must be smaller than the available shared bandwidth. Then we have:
	\begin{align}
	\label{eq:3}
	\sum\limits_{i=1}^{K} p_i \leq b.
	\end{align}
	
	\subsubsection{Model Constraint}
	Since only one model can be selected simultaneously for video stream $v_i$ on the edge device, thus:
	\begin{align}
	\label{eq:4}
	\sum\limits_{j=1}^{N} x_{ij}= 1   \quad i=1,2,3\ldots,K.
	\end{align}

	Our objective is to minimize the latency of multiple video streams with heterogeneous QoS requirements while maximizing vision detection accuracy. Then the objective function is written as:
	\begin{align}
	\label{eq:objective}
	J(x_{ij}, r_i) = \sum\limits_{i=1}^K  \frac{1}{q_i} (\sum\limits_{j=1}^N x_{ij}l_{ij} - \omega \sum\limits_{j=1}^N x_{ij}a_{ij}  )  
	\end{align}
	
	The problem is:
	\begin{subequations}\label{eq:p1}
		\begin{align}
		\mathcal{P}1: 	& \min_{x_{ij}, r_i}   J(x_{ij}, r_i)  
		\end{align} 
		\begin{alignat}{2}
		&	\text{s.t.}   &   \eqref{eq:1} - \eqref{eq:4} \\
		&	\text{vars}   & \quad x_{ij} \in \{0, 1\}  
		\end{alignat}
	\end{subequations}
	The weight parameter $\omega$ measures the relative importance of latency towards the accuracy metric.
	
	\begin{algorithm}[!t] 
		\caption{The Proposed CANS Algorithm} 
		\label{alg:Framwork} 
		\begin{algorithmic}[1] 
			\REQUIRE ~~\\ 
			QoS demand $q_i$ and Network latency $L_i$ of the $i$-th video; network bandwidth $b$; \\
			Latency model $l_{ij}$ and accuracy model $a_{ij}$;
			\ENSURE ~~\\ 
			Frame resolution $r_i^*$; \\
			Decision of CNN model for the $i$-th video $x_{ij}$;
			\STATE Initialize $r_i$, $x_{ij}$;
			\STATE $u_{\text{max}} = \sum\limits_{i=1}^K \frac{1}{q_i} (\sum\limits_{j=1}^N x_{ij}l_{ij} - \omega \sum\limits_{j=1}^N x_{ij}a_{ij}  ) $
			\STATE \textbf{If} the bandwidth fluctuates over a threshold $b_{0}$ \textbf{then}:
			\STATE \quad Running Reconfiguration;
			\STATE \quad \textbf{For} $j=1$ to $N$:  
			\STATE	\quad \quad $r_i = \arg \min\limits_{x_{ij}, r_i} \sum\limits_{i=1}^K \frac{1}{q_i} (\sum\limits_{j=1}^N x_{ij}l_{ij} - \omega \sum\limits_{j=1}^N x_{ij}a_{ij}  )$;
			\STATE \quad \quad \textbf{If}  $\sum\limits_{j=1}^N x_{ij} l_{ij}^{\text{CNN}}\leq \frac{1}{f_i} $ \AND $\sum\limits_{i=1}^{K} p_i \leq b$ \AND $\sum\limits_{j=1}^{N} l_{ij} x_{ij} \leq \min\limits_{i} L_i $ \textbf{then}:
			\STATE  \quad \quad $u_{\text{max}} \leftarrow u $;
			\STATE \quad \quad $r_i^* \leftarrow r_i, x_{ij}^* \leftarrow x_{ij}$ ;
			\RETURN $r_i^*, x_{ij}^*$; 
		\end{algorithmic}
		\vspace{0pt}
	\end{algorithm}
	
	The problem $\mathcal{P}1$ is a non-linear multiple-constraint knapsack program. The multiple constraints come from the latency and bandwidth restrictions. The main difference from the general problem is that these key metrics are functions of the decision variables. Note that these functions are generally non-linear and can only be obtained empirically from experiments. 
	

	\subsection{Adaptive Camera Network Configuration Algorithm}
	The most simple solution to $\mathcal{P}1$ is a brute-force algorithm with $O(r_{\text{max}}^K \cdot N!)$ per frame. The brute-force algorithm is impractical if the configuration runs frequently or the system scale becomes too large. Thus, we design an efficient algorithm to solve $\mathcal{P}1$. Algorithm \ref{alg:Framwork} lists the procedure of our CANS algorithm. The computational complexity of the proposed algorithm is reduced to $O(r_{\text{max}} \cdot NK)$. 
	
	\textit{Step 1:} The frame resolution $r_i$ and the model decision $x_{ij}$ are initialized with bandwidth constraints. 
	
	\textit{Step 2:} As the program runs and the bandwidth varies with time, the edge device monitors the bandwidth variation to determine when running reconfiguration. Suppose the network resources fluctuate over a predetermined threshold, \emph{e.g.,} 10\%. In that case, the CANS restarts to reconfigure a proper video profile for each video stream $v_i$. 
	\begin{align}
	r_i = \arg \min\limits_{x_{ij}, r_i} \sum\limits_{i=1}^K \frac{1}{q_i} (\sum\limits_{j=1}^N x_{ij}l_{ij} - \omega \sum\limits_{j=1}^N x_{ij}a_{ij}  )
	\end{align}
	Besides, the CANS also selects a proper detection model on the edge device. 
	
	\textit{Step 3:} If the reconfiguration starts, the proposed algorithm is operated once on the edge device to re-allocate computing resources and network resources for each video stream. 
	The configuration profiles are sent back to each camera and sent to the edge device for model selection.

	Moreover, we conduct three additional modes of the proposed system as follows:
	\begin{itemize}
		\item \textbf{Accuracy-optimal} does not configure as bandwidth varies. All video streams neglect the latency constraints and select the most expensive configuration knobs.
		\item \textbf{Delay-optimal} aims to minimize the overall latency of all video streams and ignores the accuracy guarantee.
		\item \textbf{Delay-chronic} satisfies long-term service latency by imposing a hard latency constraint, short of adaptation, and purely myopic. 
		\item \textbf{CANS} considers the tradeoff between accuracy and latency under network dynamics. Moreover, the reconfiguration is executed only when the network resources change drastically, saving computing complexity and configuration costs.  
	\end{itemize} 
	The performance comparison results are shown in Section IV.

	\subsection{Accuracy Metrics}
	The accuracy of vision detection is measured by the intersection over union (IOU) metric:
	\begin{equation}
	IOU = \frac{\mathcal{B}_1 \cap \mathcal{B}_2}{\mathcal{B}_1 \cup \mathcal{B}_2},
	\end{equation}
	where $\mathcal{B}_1$ and $\mathcal{B}_2$ are the bounding boxes of two objects. If the IOU between the detected bounding box and the ground truth is over than a threshold $IOU_{\text{min}}$ (\emph{e.g.,} 0.7), the detected object is regarded as a true positive. The overall accuracy of a video is calculated by the average F1 score of all video frames.

	\begin{figure}[!t]
		\centerline{\includegraphics[width=3cm]{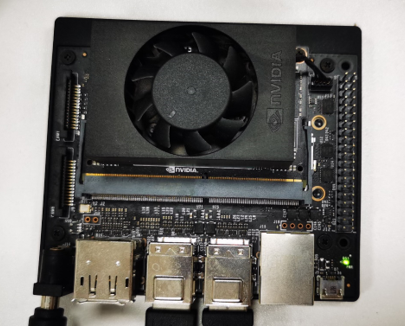}}
		\caption{The vision detection models are implemented on the edge device, an NVIDIA Jetson Xavier NX.}
		\label{fig:jetson}
		\vspace{-10pt}
	\end{figure}
	
	\begin{figure}[!t]
		\centerline{\includegraphics[width=7cm]{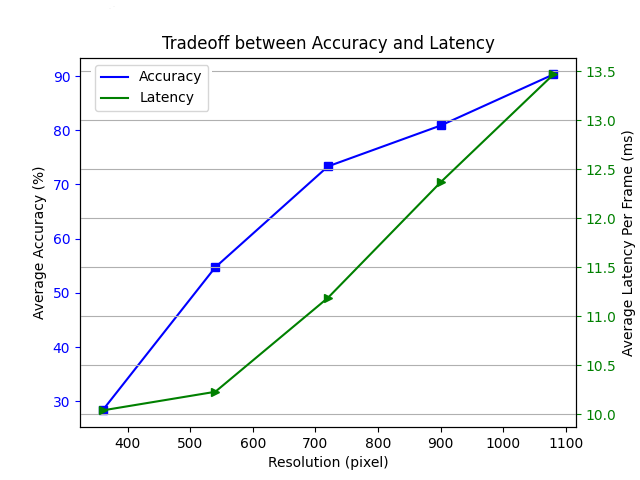}}
		\caption{The tradeoff between latency and accuracy with respect to the input image resolution of a video stream.}
		\label{fig: tradeoff}
		\vspace{-15pt}
	\end{figure}

	\section{Performance Evaluation}
	In this section, the performance of the proposed CANS is evaluated on real-world surveillance datasets. The dataset is the video data of a real-world surveillance dataset, MOT Challenge 2020 \cite{denforfer2020mot}. Extensive simulation results demonstrate the effectiveness of the proposed CANS method. 
	The edge device is implemented by an NVIDIA Jetson Xavier NX platform with three vision detection models, including SSD with MobileNet-v1 \cite{howard2017mobilenet}, SSD with MobileNet-v2 \cite{sandler2018mobilenetv2}, and SSD with InceptionNet \cite{szegedy2017inception}. 
	The input image resolutions are 360p, 540p, 720p, 900p and 1080p, respectively. The varying span of network bandwidth is from 20 Mbps to 100 Mbps. For each detection model, we profile its accuracy model with respect to the input frame resolution in real experiments. 
	$ L_{\text{max}} = \min\limits_{i} L_i,  i \in \{1,2,3\ldots,K\} $ in constraint \eqref{eq:1} is set to 80 ms. The threshold of bandwidth variation is 10\%. 
	Moreover, we set $\alpha=8$, $\omega=6$ in Eq. \eqref{eq:objective} and $IOU_{\text{min}}=0.7$.
	
	\begin{figure*}[!t]
		\centering
		\subfloat[Objective Function]{\includegraphics[width=2.3in]{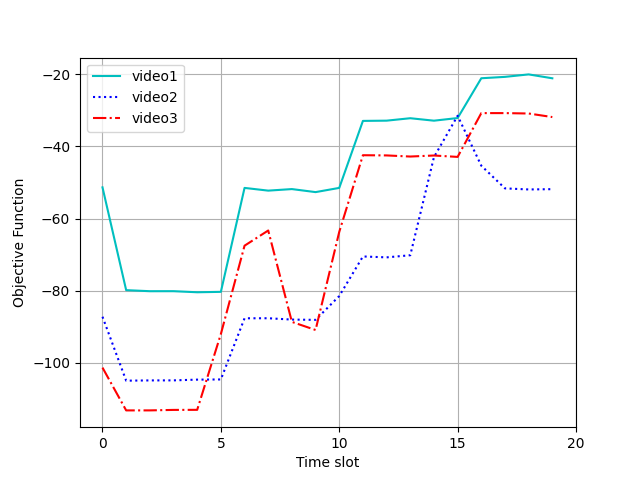}%
			\label{fig-obj}}
		\hfil
		\subfloat[Resolution]{\includegraphics[width=2.3in]{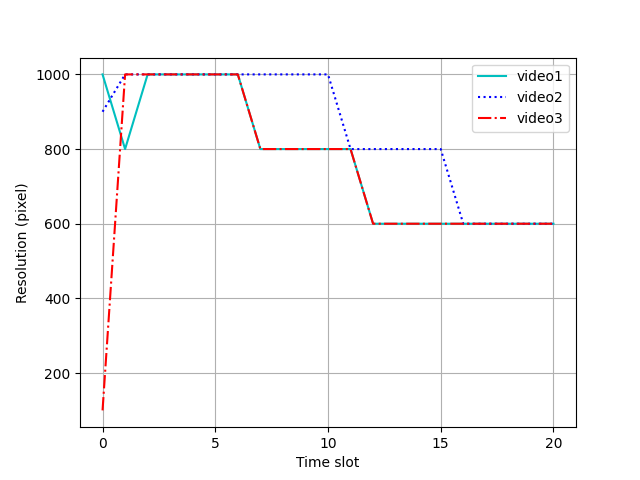}%
			\label{fig-resolu}}
		\hfil
		\subfloat[Detection Model]{\includegraphics[width=2.3in]{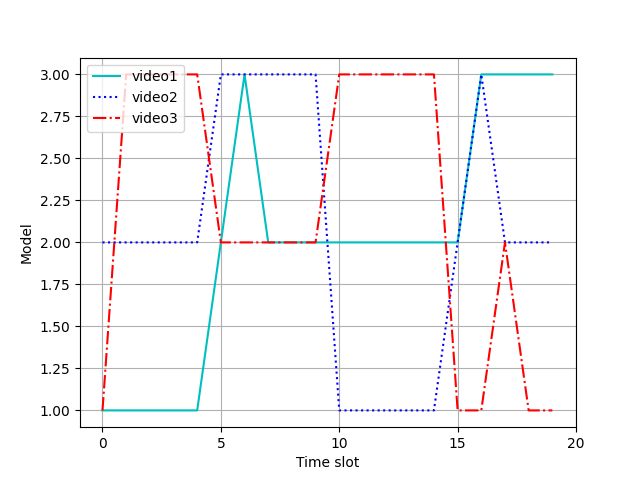}%
			\label{fig-model}}
		\caption{A running example of the proposed CANS method which adapts to network dynamics.}
		\label{fig:runningexample}
		\vspace{-15pt}
	\end{figure*}

	\begin{figure*}[htbp]
		\begin{minipage}[t]{0.33\linewidth}
			\centering
			\includegraphics[width=2.35in]{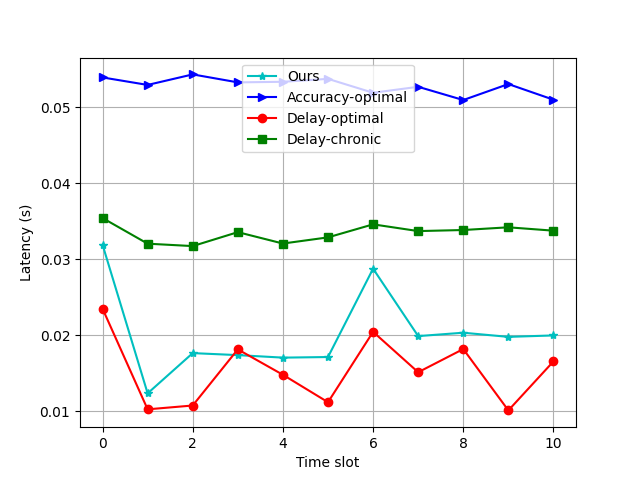}
			\caption{Latency comparison with three methods.}
			\label{fig:lat}
		\end{minipage}
		\begin{minipage}[t]{0.33\linewidth}
			\centering
			\includegraphics[width=2.1in]{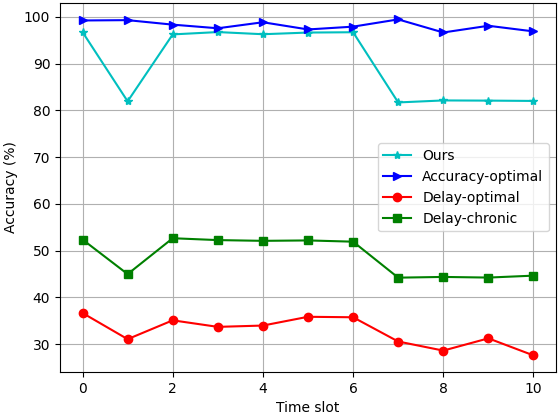}
			\caption{Accuracy comparison with three methods.}
			\label{fig:acc}
		\end{minipage}
		\begin{minipage}[t]{0.33\linewidth}
			\centering
			\includegraphics[width=2.3in]{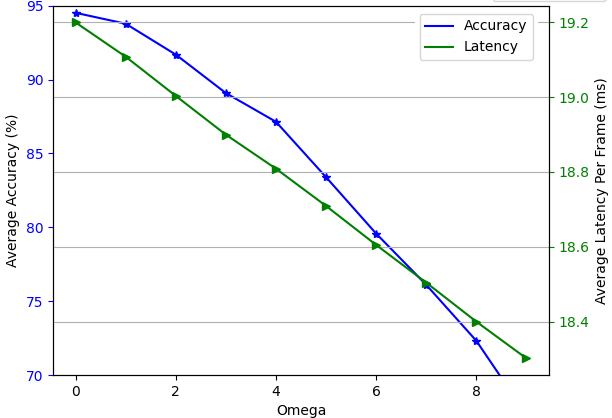}  
			\caption{The impact of $\omega$ on the latency and accuracy.}
			\label{fig:omega}
		\end{minipage}%
		\vspace{-15pt}
	\end{figure*}

	\subsection{Parametrizing Accuracy}

	We implement three object detectors (\emph{i.e.}, SSD with MobileNet-v1 \cite{howard2017mobilenet}, SSD with MobileNet-v2 \cite{sandler2018mobilenetv2}, and SSD with InceptionNet \cite{szegedy2017inception}) on NVIDIA Jetson Xavier NX2 to execute detection on real-world surveillance videos, as shown in Fig. \ref{fig:jetson}. Since manual annotation is laborious, we assume the ground-truth detection results are the detections with the most expensive configuration (\emph{golden} configuration). The accuracy is computed by comparing the detected objects of the current configuration and those of the \emph{golden} configuration. A video's accuracy is measured by the F1 score (the harmonic mean of precision and recall). 

	The blue line in Fig. \ref{fig: tradeoff} shows the results of the relationship between accuracy and input image resolution of SSD with MobileNet-v1 \cite{howard2017mobilenet}. We vary the image resolution from 360p to 1080p. 
	The first observation is that the accuracy becomes higher when the image resolution becomes larger. The performance gain decreases gradually with the increase of resolution. Thus, the relationship between accuracy and the input image resolution can be formulated as a convex function. The curve is fitted as $a=-0.0002r^2 + 0.3316r - 71.034$ by a least square method with less than 0.02 mean square error. 
	Similarly, the accuracy models of the other detection model could be derived.

	\subsection{System Adaptation Under Network Dynamics}
	Fig. \ref{fig:runningexample} illustrates how the proposed CANS adapts to bandwidth variation. In this example, three cameras connect to the same edge device with a shared channel with a randomly initialized configuration. 
	The video content changes with time slots vary. For example, occasionally the bandwidth changes drastically at time slots 5, 10, 15; then, all video streams are configured with lower video resolution to reduce bandwidth consumption. Concretely, video streams 2 and 3 switch to CNN models 3 and 2 for less accuracy loss at time slot 5, respectively.  
	
	\begin{figure*}[htbp]
		\begin{minipage}[t]{0.33\linewidth}
			\centering
			\includegraphics[width=2.3in]{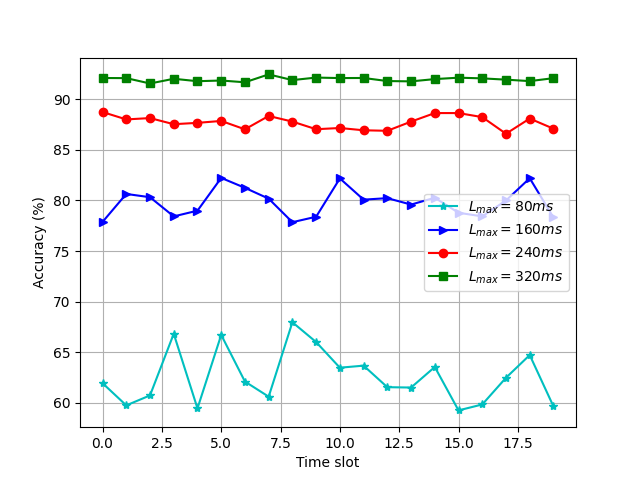}
			\caption{The impact of $L_{\text{max}}$ on the accuracy.}
			\label{fig:lmax}
		\end{minipage}%
		\begin{minipage}[t]{0.33\linewidth}
			\centering
			\includegraphics[width=2.3in]{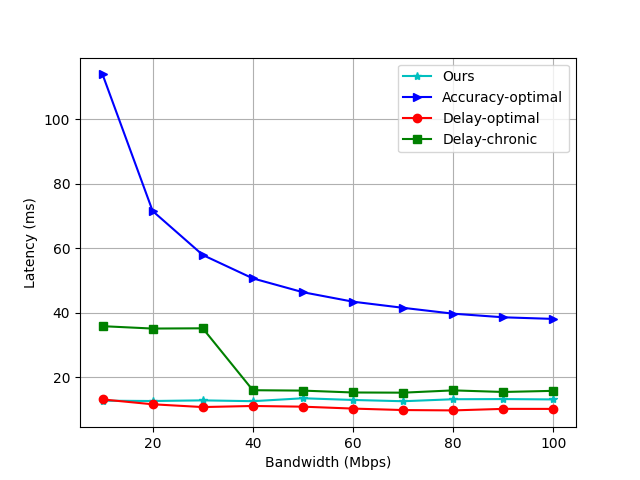}
			\caption{Latency comparison with three methods.}
			\label{fig:lat-b}
		\end{minipage}
		\begin{minipage}[t]{0.33\linewidth}
			\centering
			\includegraphics[width=2.3in]{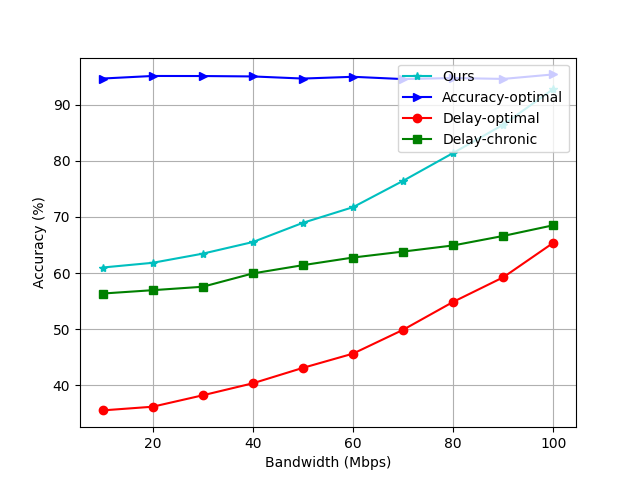}
			\caption{Accuracy comparison with three methods.}
			\label{fig:acc-b}
		\end{minipage}
		\vspace{-15pt}
	\end{figure*} 
	
	\subsection{Algorithm Study}
	The average system latency and accuracy of the proposed CANS method are shown in Fig. \ref{fig:lat} and Fig. \ref{fig:acc}, respectively. The bandwidth is 100 Mbps. The results are compared with accuracy-optimal, delay-optimal, and delay-chronic. 
	The accuracy-optimal uses the largest resolution to maintain the highest accuracy, nearly 95\%. Thus, it does not know how to choose the proper configuration knobs and has the most expensive latency cost as expected. 
	Contrastly, the delay-optimal performs the lowest latency by selecting the smallest resolution of input video streams. As a result, the delay-optimal has a significant sacrifice in the average analytics accuracy. 
	The delay-chronic is slightly more intelligent than accuracy-optimal and delay-optimal due to a hard latency constraint. This constraint makes delay-chronic meet the long-term latency requirement. The latency of the delay-chronic is 33 ms, and the analytics accuracy is 48\%. However, the delay-chronic is less adaptive, and both the accuracy and the latency metrics are inferior to the proposed CANS method. 
	
	The average accuracy of the proposed CANS achieves 92\%, performing 44\% higher than the delay-chronic. Note that the accuracy of our CANS drops only 8\% comparing with the accuracy-optimal. Thus, the comparison results report that the proposed CANS obtains satisfactory accuracy with only a marginal accuracy loss. 
	Meanwhile, the average latency of the proposed CANS is 13 ms. It is approximately 12 ms less than the delay-chronic mode's latency. Comparing with the delay-optimal, the latency of our CANS has only a 5 ms increase. This comparison demonstrates that our CANS is able to leverage varying network conditions and provides low response latency of video surveillance in edge-enabled IIoT.

	\subsection{Parametric Sensitivity Analysis}
	
	\subsubsection{Impacts of Parameters}
	The impact of $\omega$ on the proposed algorithm is tested, and Fig. \ref{fig:omega} shows the results. It is observed that when increasing $\omega$ from 1 to 4, the algorithm obtains 0.3 ms latency reduction with 8\% accuracy loss. Larger $\omega$ brings more concentration on latency, leading to lower latency but smaller accuracy. Oppositely, a smaller $\omega$ results in better accuracy but higher latency. Thus, balancing the tradeoff between accuracy and latency is challenging for algorithm design. 
	In the simulation, we set $\omega=6$. 
	
	Moreover, the impact of $ L_{\text{max}} = \min\limits_{i} L_i$, $ i \in \{1,2, \ldots,K\} $ is also evaluated. Fig. \ref{fig:lmax} illustrates the accuracy of CANS with respect to different $L_{\text{max}}$. The first observation is that a larger $L_{\text{max}}$ (\emph{i.e.}, a looser latency constraint) brings a higher analytics accuracy. With time varies, the accuracy has a fluctuation due to video content changes and dynamic bandwidth. 
	A small $L_{\text{max}}$ means the latency constraint is more strict and can be easily breached. Thus, occasionally an evident sacrifice of accuracy may meet the latency constraint. Contrastly, a larger $L_{\text{max}}$ leads to a smaller fluctuation of accuracy.

	\subsubsection{Impact of Bandwidth}
	Moreover, we also report the performance of the proposed CANS method and the other three methods with varying bandwidths. Fig. \ref{fig:lat-b} and Fig. \ref{fig:acc-b} show the results on the latency and accuracy comparison. 
	Our CANS achieves 92\% accuracy with 13 ms latency on average at 100 Mbps. 
	It is observed that with more available bandwidth resources, all methods except the accuracy-optimal tend to have a more considerable accuracy because more bandwidth resources are able to support more expensive configuration knobs. 
	The average latency of our CANS and the delay-chronic are bounded as bandwidths increase. The latency gap between CANS and the delay-chronic is significant when bandwidth is small. Then the latency gap becomes smaller as bandwidth increases. 
	On the contrary, the accuracy-optimal and the delay-optimal have a drastic reduction in latency with increasing bandwidth.

	\section{Conclusion}
	This paper proposes an adaptive camera network self-configuration method (CANS) for multiple video streams of heterogeneous QoS in edge-enabled IIoT. An optimization problem is formulated to reduce the total latency of video streams while maximizing the accuracy metric, subject to latency constraints and bandwidth conditions. An effective algorithm is present to solve the optimization problem with varying network bandwidths. 
	Simulation results validate the effectiveness of the proposed CANS method and its adaptation to network dynamics. 
	

	\bibliographystyle{IEEEtran}
	\bibliography{IEEEexample}

\end{document}